\documentclass[conference]{IEEEtran}
\IEEEoverridecommandlockouts
\usepackage{cite}
\usepackage{amsmath,amssymb,amsfonts}
\usepackage{algorithm}
\usepackage{algpseudocode}
\usepackage{graphicx}
\usepackage{textcomp}
\usepackage{xcolor}

\usepackage{multirow}
\usepackage{makecell}
\usepackage{tikz}
\usepackage{pgfplots}
\usepackage{stfloats}

\def\BibTeX{{\rm B\kern-.05em{\sc i\kern-.025em b}\kern-.08em
    T\kern-.1667em\lower.7ex\hbox{E}\kern-.125emX}}

\makeatletter
 \let\old@ps@headings\ps@headings
 \let\old@ps@IEEEtitlepagestyle\ps@IEEEtitlepagestyle
 \def\confheader#1{%
 \def\ps@headings{%
 \old@ps@headings%
 \def\@oddhead{\strut#1\hfill\strut}%
 \def\@evenhead{\strut#1\hfill\strut}%
 }%
 \def\ps@IEEEtitlepagestyle{%
 \old@ps@IEEEtitlepagestyle%
 \def\@oddhead{\strut#1\hfill\strut}%
 \def\@evenhead{\strut#1\hfill\strut}%
 }%
 \ps@headings%
 }
 \makeatother

\confheader{\emph{2022 IEEE NIGERCON}}

\usepackage[pscoord]{eso-pic}

\begin{document}

\title{Quantity vs. Quality of Monolingual Source Data in Automatic Text Translation: Can It Be Too Little If It Is Too Good?\\
\thanks{This work is supported by the National Information Technology Development Agency under the National Information Technology Development Fund PhD Scholarship Scheme 2018.}}

\author{\IEEEauthorblockN{1\textsuperscript{st} Idris Abdulmumin}
\IEEEauthorblockA{\textit{Department of Computer Science} \\
\textit{Ahmadu Bello University}\\
Zaria, Nigeria \\
iabdulmumin@abu.edu.ng}
\and
\IEEEauthorblockN{2\textsuperscript{nd} Bashir Shehu Galadanci}
\IEEEauthorblockA{\textit{Department of Software Engineering} \\
\textit{Bayero University}\\
Kano, Nigeria \\
bsgaladanci.se@buk.edu.ng} \\
\IEEEauthorblockN{4\textsuperscript{th} Garba Aliyu}
\IEEEauthorblockA{\textit{Department of Computer Science} \\
\textit{Ahmadu Bello University}\\
Zaria, Nigeria \\
algarba@abu.edu.ng}
\and
\IEEEauthorblockN{3\textsuperscript{rd} Shamsuddeen Hassan Muhammad}
\IEEEauthorblockA{\textit{Department of Computer Science} \\
\textit{Bayero University}\\
Kano, Nigeria \\
shmuhammad.csc@buk.edu.ng}
}

\maketitle

\begin{abstract}
Monolingual data, being readily available in large quantities, has been used to upscale the scarcely available parallel data in order to train better models for automatic translation. Self-learning, where a model is made to learn from its output, is one approach of exploiting such data. However, it has been shown that too much of this data can be detrimental to the performance of the model if the available parallel data is comparatively extremely low. In this study, we investigate whether the monolingual data can also be too little and if this reduction, based on quality, have a any affect on the performance of the translation model. Experiments have shown that on English-German low resource NMT, it is often better to select only the most useful additional data---based on quality or closeness to the domain of the test data---than utilizing all of the available data.
\end{abstract}

\begin{IEEEkeywords}
self-learning, domain adaptation, quality estimation, machine translation, natural language processing
\end{IEEEkeywords}

\section{Introduction}
Automatic text translation has enabled effective communication and delivery of contents in foreign languages to many communities, especially where local content is scarce. The problem, though, is that many language pairs do not have an adequate amount of data for training translation systems \cite{Sennrich2019}. To solve this problem, the monolingual data, which is almost always available in abundance, has been used to enable the training of more qualitative systems for translation. Example of methods that adopt the monolingual data include self-learning \cite{Ueffing2006}, back-translation \cite{Sennrich2016a}, a hybrid of the two \cite{Abdulmumin2021hybrid} and language modelling \cite{Gulcehre2017}.

The self-learning approach has been one of the most successful and easy-to-implement methods for improving the performances of automatic translation systems using the monolingual data \cite{Specia2018}. The method involves using a new or existing model to translate a huge amount of data in the source language, and then using the resulting parallel data to improve the quality of the base model. However, Gasc{\'{o}} \emph{et al.} \cite{Gasco2012} demonstrated that more training data does not always result in a better model. To achieve the optimal performance, the approach has been shown to require either skipping training on the machine-generated side of the additional data \cite{Zhang2016} or a quality estimation \cite{Ueffing2006,Specia2018} or data selection systems \cite{Abdulmumin2021dataselection} to select only the most qualitative sentences. In the last two approaches, the additional data was reduced but the performance increased.

In this study, we hybridized data selection and quality estimation to further reduce the quantity of the additional data while maintaining only the best-quality sentences during self-learning. Using our approach, we reduced the monolingual data by about 94\% and still obtained a better performance than using all of the available data. Since less data means less training and less storage, our approach is also more efficient than using large monolingual data. Thus, in this study, we make the following contributions:
\renewcommand{\labelitemi}{\textbullet}
\begin{itemize}
\item to extract the most qualitative and in-domain additional data for self-learning, we hybridized quality estimation with domain-aware data selection;
\item we demonstrated that the additional training data in self-learning can be reduced to one-sixteenth of the available data while maintaining the same performance as using a larger subset of the data and a better performance than using all of the available data;
\item experimental results on English-German low resource translation indicate that the method is applicable on, and can be generalized to, other low resource languages.
\end{itemize}

The paper is organized as follows: Review of related literature is given Section~\ref{sec:literature}, the proposed approach in Section~\ref{sec:methods}, experimental set-up and dataset are presented in Section~\ref{sec:experiment}, the experimental results are discussed in Section~\ref{sec:results} and lastly, the study is concluded in Section~\ref{sec:conclude}.

\section{Related Works}
\label{sec:literature}

Self-learning has been used in machine translation to improve quality of the translation models. Ueffing \cite{Ueffing2006} utilized the method to improve a phrase-based statistical translation system. The authors leveraged an existing system to generate translations of a fresh collection of monolingual source data. Following that, the confidence score of each of the translated sentences is evaluated, and translations that are deemed reliable are extracted and used as new training data to improve the existing system based on these scores. \cite{Zhang2016} then proposed freezing the parameters of the decoder in NMT when training on the synthetic data. \cite{Specia2018} proposed an iterative implementation of the approach in \cite{Ueffing2006} where all of the available monolingual data are utilized.

He \emph{et al.} \cite{He2020} concluded that introducing noise into the additional improves performance. In a similar setting, although in back-translation, Caswell \emph{et al.} \cite{Caswell2019} found that the noising only indicates to the model that a particular instance of the training data is either authentic or synthetic, resulting in the improvement of the model's performance. Instead, Abdulmumin \emph{et al.} \cite{Abdulmumin2021enhanced} used pre-training on the additional data and fine-tuning on the authentic data to enable the model to differentiate between the two data. This was shown to improve its performance. Abdulmumin \emph{et al.} \cite{Abdulmumin2021hybrid} used the implementation by Ueffing \cite{Ueffing2006} and Specia and Shah \cite{Specia2018} to improve back-translation. Abdulmumin \emph{et al.} \cite{Abdulmumin2021dataselection} showed that data selection can be used in languages where quality estimation cannot be done due to the absence of such systems or the data to train them.

\section{Methodology}
\label{sec:methods}
In this section, we described the neural architecture used in building the models and the proposed self-learning approach.

\subsection{Neural Machine Translation (NMT)}

The NMT is simple form of machine translation architecture than can learn to translate between the input text in the source language to sentences in the target language from the available parallel data between these two languages \cite{Bahdanau2015}. Given an input sequence \(X=(x_1,...,x_{T_x})\) and previously translated words \((y_1,...,y_{i-1})\), the probability of the next word \(y_i\) is given by \begin{equation} p(y_i|y_1,...,y_{i-1},X) = g(y_{i-1},s_i,c_i) \end{equation}, where \(s_i\) is the decoder hidden state for time step \(i\) and is computed as \begin{equation} s_i = f(s_{i-1},y_{i-1},c_i) \end{equation} Here, \(f \) and \(g\) are nonlinear transform functions, which can be implemented as long short-term memory (LSTM) network \cite{Hochreiter1997} in recurrent neural machine translation (RNMT), and \(c_i\) is a distinct context vector at time step \(i\), which is calculated as a weighted sum of the input annotations, \(h_j\), which is a concatenation of the forward and backward hidden states \(\overrightarrow{h_j}\) and \(\overleftarrow{h_j}\) respectively.
\begin{equation}
c_i=\sum_{j=1}^{T_x} a_{i,j}h_j
\end{equation}
where \(h_j\) is the annotation of \(x_j\) calculated by a bidirectional Recurrent Neural Network. The weight \(a_{i,j}\) for \(h_j\) is calculated as
\begin{equation}
a_{i,j} = \frac{\exp{e_{i,j}}}{\sum_{t=1}^{T_x} \exp{e_{i,t}}}
\end{equation}
and
\begin{equation}
e_{i,j} = v_a\tanh(Ws_{i-1}+Uh_j)
\end{equation}
where \(v_a\) is the weight vector, \(W\) and \(U\) are the weight matrices.

\subsection{Proposed Self-Learning with Data Selection and Quality Estimation}
The proposed approach is presented in Algorithm~\ref{algo-sldsqebt}. Given a set of parallel training data and the monolingual source sentences of sizes \(U\) and \(V\): \(D^p =\{(x^{(a)}, y^{(a)})\}_{a=1}^A\) and \(X =\{(x^{(b)})\}_{b=1}^B\) respectively, the monolingual sentences were ranked according to how closely their domain matched that of the test set using a data selection system. Then, the top \(n\) sentences from this sorted sentences were selected. After that, the authentic parallel data, \(D^p\), was used to train a translation model, \(M_{x \rightarrow y}\). The model was the utilized to produce synthetic parallel data: \(D^\prime =\{(x^{(n)}, y^{(n)})\}_{n=1}^N\) by translating the selected monolingual source data, \(X^N\), where \(N\) is the size of the data. The best \(m\) synthetic sentences were then chosen using a quality estimation system. The resulting synthetic data was then used to train an improved model through pre-training and fine-tuning on the synthetic and authentic data respectively.

\begin{algorithm*}[t!]
\renewcommand{\arraystretch}{1.3}
\caption{Proposed Self-Learning with Domain-aware Data Selection and Quality Estimation}
\label{algo-sldsqebt}
\textbf{Input}: Parallel data, \(D^p =\{(x^{(a)}, y^{(a)})\}_{a=1}^A\), Monolingual source data, \(X =\{(x^{(b)})\}_{b=1}^B\), \(n = \), the Test set, \(T = \{(x^{(c)}, y^{(c)})\}_{c=1}^C\), number of required monolingual source data, and \(m = \) number of required synthetic data, with \(n > m\)
\textbf{Output}: Improved translation model, \(M_{x \rightarrow y}\)
\begin{algorithmic}[1]
\Procedure{Domain-aware Data Selection}{}
    \State Sort \(X\), from nearest to farthest, depending on the closeness of the domain of \(x^{(b)} \in X\) to that of \(x \in T\), using a Data Selection algorithm;
    \State Let \(X^n = \) the nearest \(n\) monolingual source sentences;
\EndProcedure
\Procedure{Self-Learning with Quality Estimation}{}
    \State Train a translation model, \(M_{x \rightarrow y}\) on parallel data \(D^p\);
    \State Use \(M_{x \rightarrow y}\) to create the synthetic data, \(D^\prime = \{(x^{(n)}, y^{(n)})\}_{n=1}^N \forall x \in X^N\);
    \State Using a \(QE\) system, select best \(m\) synthetic data, \(D^m\), from \(D^\prime\);
    \State Train an enhanced translation model \(M_{x \rightarrow y}^+\) on parallel data \(D^p\) and \(D^m\);
\EndProcedure
\end{algorithmic}
\end{algorithm*}

\section{Experimental Set-up}
\label{sec:experiment}

In this section, we present the experimental set-up and the training and evaluation datasets.

\subsection{Data}
\label{sec:data}

The work is implemented using the IWSLT 2014 German-English (De-En) shared translation task data \cite{Cettolo2014}. We used the \cite{Ranzato2016} clean-up and splitting script to generate the train, development and evaluation sets. From the pre-processed \cite{Luong2015} WMT 2014 English-German (En-De) translation task \cite{Bojar2017}, we extracted and used 400,000 random English sentences as the monolingual data. Table~\ref{tab-data} shows the statistics of the datasets.

\begin{table}[H]
\renewcommand{\arraystretch}{1.3}
\caption{Training and Evaluation Dataset}
\label{tab-data}
\begin{tabular}{|c|c|c|c|c|c|}
\hline
\multirow{2}{*}{data} & \multicolumn{3}{c|}{train} & \multirow{2}{*}{dev} & \multirow{2}{*}{test} \\ \cline{2-4}
 & sentences & \multicolumn{2}{c|}{tokens (unique)} & & \\ \hline
\multirow{2}{*}{} & \multirow{2}{*}{} & English & German & \multirow{2}{*}{} & \multirow{2}{*}{} \\ \cline{3-4}
parallel & 153,348 & \makecell[tc]{2,706,255\\ (54,169)} & \makecell[tc]{3,311,508\\ (25,615)} & 6,970 & 6,750 \\ \hline
monolingual & 400,000 & \multicolumn{2}{c|}{9,918,380 (266,640)} & -- & -- \\ \hline
\end{tabular}
\end{table}

\begin{table*}[b!]
\renewcommand{\arraystretch}{1.3}
\centering
\caption{Translation performances with and without domain-aware data selection and synthetic data quality estimation for German-English NMT. KEY: P = pre-train and F = fine-tune.}
\label{slqeds-tab1}
\begin{tabular}{|c|c|c|c|c|c|c|c|c|c|c|c|c|c|}
\hline
\multirow{3}{*}{checkpoint} & \multirow{3}{*}{baseline} & \multicolumn{12}{c|}{self-training with \emph{synth} (no. of monolingual sentences)} \\ \cline{3-14}

& & \multicolumn{2}{c|}{+ \emph{all (400k)} \cite{Abdulmumin2021enhanced}} & \multicolumn{2}{c|}{+ \emph{qe (133k)} \cite{Abdulmumin2021hybrid}} & \multicolumn{2}{c|}{+ \emph{ds (133k)} \cite{Abdulmumin2021dataselection}} & \multicolumn{2}{c|}{+ \emph{ds-qe (100k)}} & \multicolumn{2}{c|}{+ \emph{ds-qe (50k)}} & \multicolumn{2}{c|}{+ \emph{ds-qe (25k)}} \\ \cline{3-14}

& & \emph{P} & \emph{F} & \emph{P} & \emph{F} & \emph{P} & \emph{F} & \emph{P} & \emph{F} & \emph{P} & \emph{F} & \emph{P} & \emph{F} \\ \hline

\begin{tabular}[c]{@{}c@{}}best \\ (training step)\end{tabular} & \begin{tabular}[c]{@{}c@{}}10.03 \\ (65k) \end{tabular} & \begin{tabular}[c]{@{}c@{}}6.02 \\ (75k) \end{tabular} & \begin{tabular}[c]{@{}c@{}}20.80 \\ (125k)\end{tabular} & \begin{tabular}[c]{@{}c@{}}6.24 \\ (65k) \end{tabular} & \begin{tabular}[c]{@{}c@{}}23.22 \\ (165k)\end{tabular} & \begin{tabular}[c]{@{}c@{}}7.35 \\ (55k)\end{tabular} & \begin{tabular}[c]{@{}c@{}}23.07 \\ (140k)\end{tabular} & \begin{tabular}[c]{@{}c@{}}7.00 \\ (50k)\end{tabular} & \begin{tabular}[c]{@{}c@{}}23.01 \\ (155k)\end{tabular} & \begin{tabular}[c]{@{}c@{}} 5.14 \\ (40k)\end{tabular} & \begin{tabular}[c]{@{}c@{}} 23.09 \\ (155k)\end{tabular} & \begin{tabular}[c]{@{}c@{}} 1.91 \\ (45k)\end{tabular} & \begin{tabular}[c]{@{}c@{}} 23.06 \\ (135k)\end{tabular} \\ \hline

average & 10.25 & 6.01 & 21.31 & 6.23 & 23.66 & 7.16 & 23.34 & 6.69 & 23.42 & 5.13 & 23.78 & 1.58 & 23.42 \\ \hline
\end{tabular}
\end{table*}

\subsection{Set-up}
\label{sec:setup}
In this work, we used a similar set-up to Abdulmumin \emph{et al.} \cite{Abdulmumin2021dataselection}. The OpenNMT-tf \cite{Klein2017} framework, implemented using TensorFlow \cite{Abadi2015}, was used to build and train the models. The model configuration used is the pre-configured NMTSmallV1 -- a 2-layer unidirectional LSTM encoder-decoder with 512 hidden units each and Luong attention \cite{Luong2015}, Adam optimizer \cite{Kingma2015}, a batch size of 64, a static learning rate of 0.0002 and a dropout probability of 0.3.

We learned BPE \cite{Sennrich2016b} on the training data with 10,000 merge operations before applying it to the train, development, and evaluation data for data pre-processing. Following that, we created a training data vocabulary. For each of the source and target vocabulary sizes, we utilized 50,000 unique tokens. Each of the self-learning approaches were trained using the \emph{"joint BPE"} \cite{Kocmi2018} pre-training and fine-tuning approach \cite{Abdulmumin2021enhanced}. For estimating the quality of the synthetic sentences, we used the pre-trained Predictor-Estimator \cite{Kim2017} model in OpenKiwi’s \cite{Kepler2019} tutorial. The model, a Recurrent Neural Network architecture trained using stack propagation \cite{Zhang2016a}, utilizes multi-level task learning for synthetic data quality estimation. For data selection, we used the frequency decay algorithm \cite{Bicici2011}. This algorithm compares the \(n\)-grams of the source-side of the test set and the monolingual source sentences to determine in-domain sentences in the monolingual data.

For the evaluation, we used the BLEU metric \cite{Papineni2002}. The models were evaluated after every 5,000 training steps on the development set during training and is stopped when less than 0.2 BLEU is the maximum average improvement observed after the evaluation of four consecutive checkpoints. Afterwards, the checkpoints are evaluated on the test set. The best models were obtained after averaging the last 8 checkpoints. Finally, the proposed approach is evaluated against self-learning with data selection (DS) \cite{Abdulmumin2021dataselection} and quality estimation (QE) \cite{Abdulmumin2021hybrid}, and without any of these approaches \cite{Abdulmumin2021enhanced}.

\section{Experiment and Results}
\label{sec:results}
Firstly, a baseline model was trained on the available parallel data. Secondly, the domain-aware data selection method described above was used to select the best one-third (133,317) of the available monolingual sentences. Afterwards, this data was translated using the baseline model. The QE system was then used to select 100,000, 50,000 and 25,000 synthetic sentences that are estimated to be the most qualitative. The additional data were used during pre-training and the authentic parallel data to fine-tune the models. Cumulatively, the models trained for 180,000, 160,000 and 150,000 steps respectively before they stopped. During pre-training, the best model reached the climax performance of 7 BLEU on 133,317 synthetic data at the 50,000$^{th}$ step. The fine-tuned best model reached a performance of 23.78 on only 50,000 sentences after checkpoint averaging.

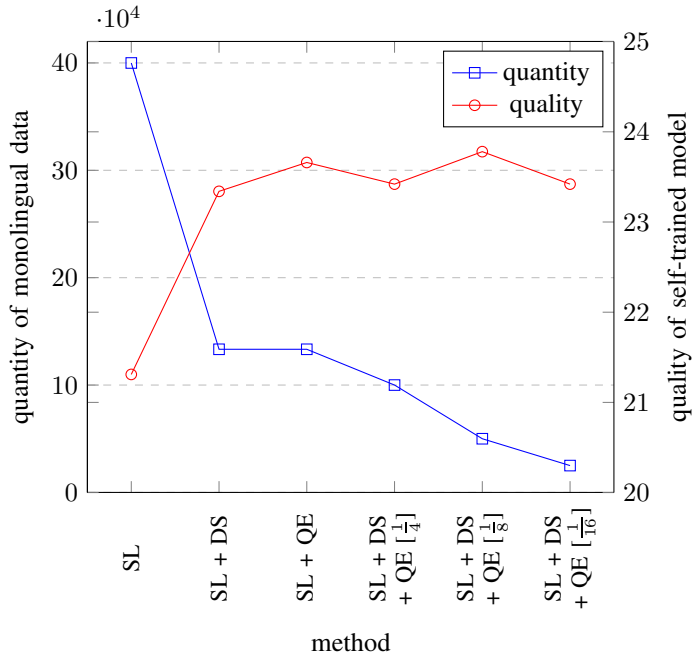
\begin{figure}[t!]
\centering
\begin{tikzpicture}

\pgfplotsset{
   width=7cm,
   height=6cm
}

\begin{axis}[
    axis y line*=left,
    ymajorgrids=true,
    grid style=dashed,
    scale only axis,
    scaled y ticks=base 10:-4,
    ymin=0, ymax = 420000,
    xlabel near ticks,
    ylabel near ticks,
    symbolic x coords = {
       SL,
       SL + DS,
       SL + QE,
       SL + DS + QE [$\frac{1}{4}$],
       SL + DS + QE [$\frac{1}{8}$],
       SL + DS + QE [$\frac{1}{16}$]
    },
    x tick label style = {
        font=\small,
        text width=1.5cm,
        align=center,
        rotate=90
    },
    xtick=data,
    xlabel={method},
    ylabel={quantity of monolingual data},
]

\addplot[
    color=blue,
    mark=square
    ]
    coordinates {({SL}, 399951) ({SL + DS}, 133317) ({SL + QE}, 133317) ({SL + DS + QE [$\frac{1}{4}$]}, 100000) ({SL + DS + QE [$\frac{1}{8}$]}, 50000) ({SL + DS + QE [$\frac{1}{16}$]}, 25000)};
    \label{plot_quantity}

\end{axis}

\begin{axis}[
    ylabel near ticks,
    yticklabel pos=right,
    axis x line=none,
    scale only axis,
    ymin=20, ymax = 25,
    symbolic x coords = {
       SL,
       SL + DS,
       SL + QE,
       SL + DS + QE [$\frac{1}{4}$],
       SL + DS + QE [$\frac{1}{8}$],
       SL + DS + QE [$\frac{1}{16}$]
    },
    ylabel={quality of self-trained model},
]

\addlegendimage{/pgfplots/refstyle=plot_quantity}\addlegendentry{quantity}

\addplot[
    color=red,
    mark=o
    ]
    coordinates {({SL}, 21.31) ({SL + DS}, 23.34) ({SL + QE}, 23.66) ({SL + DS + QE [$\frac{1}{4}$]}, 23.42) ({SL + DS + QE [$\frac{1}{8}$]}, 23.78) ({SL + DS + QE [$\frac{1}{16}$]}, 23.42)};
    \addlegendentry{quality}

\end{axis}

\end{tikzpicture}
\caption{Quantity vs Quality: Showing the quality of machine translation improving despite decreasing the monolingual training data.}
\label{fig:sldsqe}
\end{figure}

Table~\ref{slqeds-tab1} shows the performance of the improved models on the smaller amounts of monolingual data in comparison to the other models that were improved using only QE \cite{Abdulmumin2021hybrid} (best one-third of the synthetic data), only DS \cite{Abdulmumin2021dataselection} (closest one-third of the monolingual data) and all of the available monolingual data \cite{Abdulmumin2021enhanced}. Comparing the performance of the best model in our approach to the other models after pre-training on the smaller additional data, our model ranked $1^{st}$ on 50k monolingual data, outperforming the next best, self-learning with QE by $+0.12$ BLEU. It is also the fastest to converge, at 40k training steps during pre-training. Fig.~\ref{fig:sldsqe} shows the performances of these models in comparison to the amount of additional training data used.

\begin{table}[b]
\renewcommand{\arraystretch}{1.3}
\centering
\caption{Performances of the translation models that were trained using the various iterative self-learning approaches.}
\label{isldsqe-tab2}
\begin{tabular}{|c|c|c|c|c|}
\hline
\multirow{2}{*}{methods} & \multirow{2}{*}{iterations} & \multirow{2}{*}{\begin{tabular}[c]{@{}c@{}}amount of\\additional data\\\end{tabular}} & \multicolumn{2}{c|}{\begin{tabular}[c]{@{}c@{}}average of last \\ 8 checkpoints \end{tabular}} \\ \cline{4-5}
 & & & pre-train & fine-tune \\ \hline
 
\multirow{2}{*}{\begin{tabular}[c]{@{}r@{}}\emph{iter. sl} \cite{Abdulmumin2021enhanced}\end{tabular}}
 & $1^{st}$ & 400,000 & 19.16 & 24.06 \\ \cline{2-5}
 & $2^{nd}$ & 400,000 & 21.51 & 24.27 \\ \hline

\multirow{2}{*}{\begin{tabular}[c]{@{}r@{}}\emph{iter. sl}\\with QE \cite{Abdulmumin2021hybrid}\end{tabular}}
 & $1^{st}$ & 266,634 & 15.88 & 23.62 \\ \cline{2-5}
 & $2^{nd}$ & 399,951 & 19.83 & 24.05 \\ \hline

\multirow{2}{*}{\begin{tabular}[c]{@{}r@{}} \emph{iter. sl}\\with DS \cite{Abdulmumin2021dataselection}\end{tabular}}
 & $1^{st}$ & 266,634 & 20.54 & 24.24 \\ \cline{2-5}
 & $2^{nd}$ & 399,951 & 21.45 & 24.31 \\ \hline

\multirow{2}{*}{\begin{tabular}[c]{@{}r@{}} \emph{iter. sl}\\with DS \& QE \end{tabular}}
 & $1^{st}$ & 200,000 & 20.20 & 24.10 \\ \cline{2-5}
 & $2^{nd}$ & 300,000 & 21.42 & 24.15 \\ \hline
\end{tabular}
\end{table}

\begin{figure}[t!]
\centering
\begin{minipage}[t]{0.5\textwidth}
\resizebox{\linewidth}{!}{%

\begin{tikzpicture}

\pgfplotsset{
   width=6cm,
   height=4cm,
   legend style={at={(0.64,0.02)}, anchor=south west}
}

\begin{axis}[
    axis y line*=left,
    ymajorgrids=true,
    grid style=dashed,
    scale only axis,
    scaled y ticks=base 10:-4,
    ymin=80000, ymax = 420000,
    xlabel near ticks,
    ylabel near ticks,
    symbolic x coords = {
       SL,
       SL + DS,
       SL + QE,
       SL + DS + QE    
    },
    x tick label style = {
        font=\small,
        text width=1.5cm,
        align=center,
        rotate=0
    },
    xtick=data,
    xlabel={method},
    ylabel={quantity},
]

\addplot[
    color=blue,
    mark=square
    ]
    coordinates {({SL}, 399951) ({SL + DS}, 266634) ({SL + QE}, 266634) ({SL + DS + QE}, 200000)};
    \label{plot_quantity}

\end{axis}

\begin{axis}[
    ylabel near ticks,
    yticklabel pos=right,
    axis x line=none,
    scale only axis,
    ymin=21, ymax = 25,
    symbolic x coords = {
       SL,
       SL + DS,
       SL + QE,
       SL + DS + QE    
    },
    ylabel={quality},
]

\addlegendimage{/pgfplots/refstyle=plot_quantity}\addlegendentry{quantity}

\addplot[
    color=red,
    mark=o
    ]
    coordinates {({SL}, 24.06) ({SL + DS}, 24.24) ({SL + QE}, 23.62) ({SL + DS + QE}, 24.10)};
    \addlegendentry{quality}

\end{axis}
\end{tikzpicture}

}
\centering
~(a) $1^{st}$ Iteration
\end{minipage}%
\hfill
\begin{minipage}[t]{0.5\textwidth}
\resizebox{\linewidth}{!}{%
\raggedleft

\begin{tikzpicture}

\pgfplotsset{
   width=6cm,
   height=4cm,
   legend style={at={(0.64,0.02)}, anchor=south west}
}

\begin{axis}[
    axis y line*=left,
    ymajorgrids=true,
    grid style=dashed,
    scale only axis,
    scaled y ticks=base 10:-4,
    ymin=80000, ymax = 420000,
    xlabel near ticks,
    ylabel near ticks,
    symbolic x coords = {
       SL,
       SL + DS,
       SL + QE,
       SL + DS + QE    
    },
    x tick label style = {
        font=\small,
        text width=1.5cm,
        align=center,
        rotate=0
    },
    xtick=data,
    xlabel={method},
    ylabel={quantity},
]

\addplot[
    color=blue,
    mark=square
    ]
    coordinates {({SL}, 399951) ({SL + DS}, 399951) ({SL + QE}, 399951) ({SL + DS + QE}, 300000)};
    \label{plot_quantity}

\end{axis}

\begin{axis}[
    ylabel near ticks,
    yticklabel pos=right,
    axis x line=none,
    scale only axis,
    ymin=21, ymax = 25,
    symbolic x coords = {
       SL,
       SL + DS,
       SL + QE,
       SL + DS + QE    
    },
    ylabel={quality},
]

\addlegendimage{/pgfplots/refstyle=plot_quantity}\addlegendentry{quantity}

\addplot[
    color=red,
    mark=o
    ]
    coordinates {({SL}, 24.27) ({SL + DS}, 24.31) ({SL + QE}, 24.05) ({SL + DS + QE}, 24.15)};
    \addlegendentry{quality}

\end{axis}
\end{tikzpicture}

}
\centering
~(b) $2^{nd}$ Iteration \textcolor{white}{ddddddddddddddddddddddddddddddddddddddddddddddddddddd}
\end{minipage}

\caption{Performances on the test set, after first and second iterations, of the models that were trained using the iterative self-learning with quality estimation, iterative self-learning with data selection and the hybrid of both approaches.}
\label{fig:isldsqe}

\end{figure}
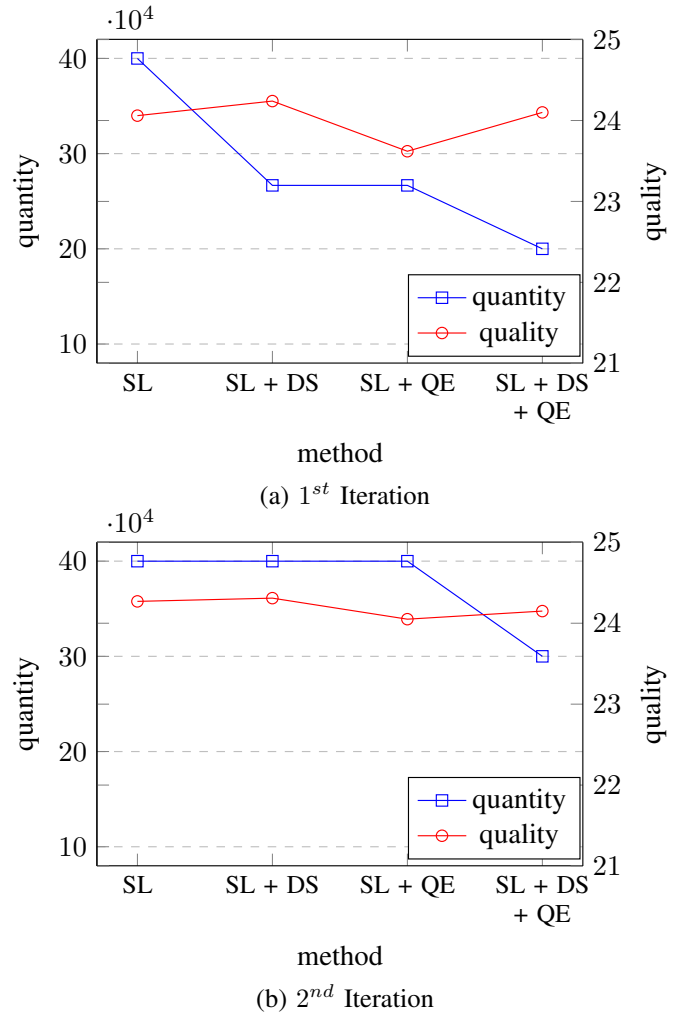

The results show that despite utilizing only one-eighth of the monolingual data, the model was still able to outperform models that were trained on one-third and all of the monolingual data. The performances of using one-third of the monolingual data in quality estimation enhanced self-learning and one-sixteenth of the data in our hybrid approach of data selection and quality estimation were also found to be similar after fine-tuning.

Furthermore, we iterated the process, taking the best of the next closest monolingual synthetic at each iteration and adding it to the additional data to self-train a more qualitative model. It can be observed from Table~\ref{isldsqe-tab2} and Fig.~\ref{fig:isldsqe} that using the same amount of 400,000 sentences throughout the iterations, the model in self-learning achieved the best performance of 24.27 BLEU. Notwithstanding, using only 200,000 and 300,000 additional data at the first and second iterations respectively resulted in a similar performance ($-0.1$ BLEU).

\section{Conclusion \& Future work}
\label{sec:conclude}
In this study, we investigated the effects of reducing the additional training data in self-learning on the performance of the translation model. We showed that if the quality of the additional data is assured, using a smaller, more-qualitative and in-domain subset of the huge amounts of the available monolingual data will produce a performance that is similar to using a larger subset of the data, and even better than using all of the available data. We showed that the additional data is never too little if it is qualitative.

At some point, though, the quantity of the additional data may be extremely insufficient to add to the knowledge of the model. While we are able to show better performance by reducing the data, it may be at the detriment of the generalizability of the model. We plan, therefore, to investigate the effects of our approach on this metric. We plan, also, to study the applicability of the approach on the similar back-translation approach and also, on rich-resourced languages.

\bibliographystyle{BibTeXtran}
\bibliography{citations}

\end{document}